\documentclass[letterpaper, 10 pt, journal, twoside]{IEEEtran}  % RA-L 

\IEEEoverridecommandlockouts                              % This command is only needed if 
\usepackage{graphics} % for pdf, bitmapped graphics files
\usepackage{epsfig} % for postscript graphics files
\usepackage{amsmath}
\usepackage{amssymb}  % assumes amsmath package installed
\usepackage{hyperref}
\usepackage{subcaption}
\usepackage[table]{xcolor}
\usepackage{collcell}
\usepackage{hhline}
\usepackage{pgf}
\usepackage{pgfplots}
\usetikzlibrary{fpu}
\usepackage{multirow}
\usepackage{booktabs,dcolumn}
\usepackage{etoolbox,siunitx}
\usepackage{xcolor}
\hypersetup{
	colorlinks,
	citecolor={blue!50!black},
	linkcolor=blue,
	filecolor=magenta,      
	urlcolor=cyan,
}
\newcolumntype{d}[1]{D{.}{.}{#1}} % "decimal" column type
\renewcommand{\ast}{{}^{\textstyle *}} % for raised "asterisks"

\def\colorModel{hsb} %You can use rgb or hsb

\pgfset{fpu/output format=fixed}
\newcommand\ColCell[1]{	
	\pgfset{fpu}
	\pgfmathparse{#1>1500?1:0}  %Threshold for changing the font color into the cells
	\ifpgfmathfloatcomparison\relax\color{white}\fi
	\pgfmathsetmacro\compA{0}      %Component R or H
	\pgfmathparse{#1<1500?0:1}
	\ifpgfmathfloatcomparison
		\pgfmathsetmacro\compB{#1/1500}
	\else
		\pgfmathsetmacro\compB{1}
	\fi
	\pgfmathsetmacro\compC{1}
	\edef\x{\noexpand\centering\noexpand\cellcolor[\colorModel]{\compA,\compB,\compC}}\x #1
	\pgfset{fpu/output format=fixed}
} 
\newcolumntype{E}{>{\collectcell\ColCell}m{0.9cm}<{\endcollectcell}}  %Cell width
\newcommand*\rot{\rotatebox{90}}

\DeclareMathOperator*{\minimize}{minimize}
\DeclareMathOperator*{\subjectto}{subject\:to}

\newif \ifdebug
\debugtrue % Set to debugtrue or debugfalse

\newcounter{ctTODO}
\newcommand{\ibf}[1]{\textit{\textbf{#1}}}
\newcommand{\trajx}{\ibf{x}}
\newcommand{\trajt}{\ibf{t}}
\newcommand{\trajX}{\ibf{X}}

\newcommand{\prob}{\mathrm{P}}

\title{Enhancing Lattice-based Motion Planning\\with Introspective Learning and Reasoning}

\author{Mattias Tiger$^{1}$ and David Bergstr{\"o}m$^{1}$ and Andreas Norrstig$^{1}$ and Fredrik Heintz$^{1}$% <-this % stops a space
	\thanks{Manuscript received: October, 15, 2020; Revised January, 18, 2021; Accepted February, 20, 2021.}
	\thanks{This paper was recommended for publication by Editor Nancy Amato upon evaluation of the Associate Editor and Reviewers' comments.}
	\thanks{
	This work was partially supported by the Wallenberg AI, Autonomous Systems and Software Program (WASP) funded by the Knut and Alice Wallenberg Foundation,
	and by grants from the National Graduate School in Computer Science (CUGS), Sweden, 
	the Excellence Center at Link{\"o}ping-Lund for Information Technology (ELLIIT), 
	the TAILOR Project funded by EU Horizon 2020 research and innovation programme GA No 952215, 
	and Knut and Alice Wallenberg Foundation (KAW 2019.0350).
	}
\thanks{$^{1}$Mattias Tiger, David Bergstr{\"o}m, Andreas Norrstig and Fredrik Heintz are with the Department of Computer and Information Science, Link{\"o}ping University, Sweden.
	{\tt\small mattias.tiger@liu.se, david.bergstrom@liu.se, fredrik.heintz@liu.se}
	}
	\thanks{Digital Object Identifier (DOI): 10.1109/LRA.2021.3068550}
}

\begin{document}
	
	\maketitle
	\thispagestyle{empty}
	\pagestyle{empty}
	
	%%%%%%%%%%%%%%%%%%%%%%%%%%%%%%%%%%%%%%%%%%%%%%%%%%%%%%%%%%%%%%%%%%%%%%%%%%%%%%%%
	\begin{abstract}
		Lattice-based motion planning is a hybrid planning method where a plan is made up of discrete actions, while simultaneously also being a physically feasible trajectory.
		The planning takes both discrete and continuous aspects into account, for example action pre-conditions and collision-free action-duration in the configuration space. Safe motion planning rely on well-calibrated safety-margins for collision checking. The trajectory tracking controller must further be able to reliably execute the motions within this safety margin for the execution to be safe.
		In this work we are concerned with introspective learning and reasoning about controller performance over time. Normal controller execution of the different actions is learned using machine learning techniques with explicit uncertainty quantification, for safe usage in safety-critical applications. By increasing the model accuracy the safety margins can be reduced while maintaining the same safety as before.
		Reasoning takes place to both verify that the learned models stays safe and to improve collision checking effectiveness in the motion planner using more accurate execution predictions with a smaller safety margin. The presented approach allows for explicit awareness of controller performance under normal circumstances, and detection of incorrect performance in abnormal circumstances. Evaluation is made on the nonlinear dynamics of a quadcopter in 3D using simulation. 
		Video: \url{https://youtu.be/F0p0FFRZwM4}
	\end{abstract}
		
	%%%%%%%%%%%%%%%%%%%%%%%%%%%%%%%%%%%%%%%%%%%%%%%%%%%%%%%%%%%%%%%%%%%%%%%%%%%%%%%%
	\section{INTRODUCTION}
	\IEEEPARstart{S}{afe} motion planning is a necessity for robots navigating in dynamic, unstructured, and human-tailored environments such as indoors or in urban settings. Operating in real dynamic environments make introspective capabilities important since situations can easily change beyond reasonable design assumptions: hardware degradation, modeling errors, software bugs as well as rare external disturbances such as extreme weather or unexpected adversarial attacks from other agents. It is important to know what normal motion execution looks like from the robot's perspective and to detect when the executed behavior become abnormal in a timely manner.
	%TODO: become -> becomes i sista meningen?
	%For the motion planning to also be useful efficient implementations must be possible.
	
	\emph{Lattice-based motion planning} is one of the most frequently used motion planning technique in real implementations for automated vehicles \cite{gonzalez2016review}.
	It works by restricting motion to a finite number of pre-computed \emph{motion primitives} which moves the robot between points on a state-lattice, and a physically feasible trajectory is found as a sequence of compatible motion primitives using graph-search. It is a technique appropriate for dynamic environments \cite{Katra2015} and a fast search for the (resolution) optimal trajectory can be performed in real-time, taking comfort, safety and vehicle constraints into consideration \cite{gonzalez2016review}.
	Lattice-based motion planning has successfully been implemented on various robots \cite{RHL-MP2018,Oliveira2018,Ljungqvist2019fieldrobotics}, with for example recent advancement on the challenge\cite{gonzalez2016review} to re-plan new collision free trajectories with multiple dynamic obstacles in real-time \cite{RHL-MP2018}.
	
	It can be incredibly challenging to specify what normal behavior looks like using formal languages such as Metric Temporal Logic (MTL) \cite{Doherty249047} or Probabilistic Signal Temporal Logic (ProbSTL) \cite{TIGER2020325}, for use in modern runtime verification frameworks to perform execution monitoring of the motion plans. Machine learning can be leveraged to complement formal safety requirements with learned models of normal action execution of for example robot manipulation tasks \cite{park2016multimodal}.
	
	\begin{figure}[t!]
		\centering
		\includegraphics[width=0.95\columnwidth]{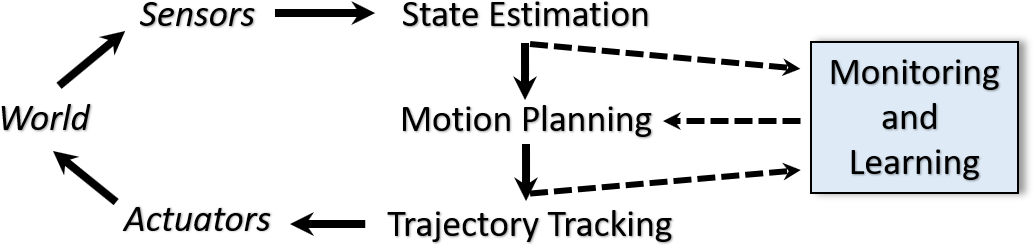}
		\caption{
			When motion execution is safety-critical, for an autonomous agent such as a robot, it is important for the robot itself to know what normal execution looks like (learning) and if the current motions are normal or abnormal (monitoring).
		}
		\label{fig:schematic-overview}
	\end{figure}
	
	Lattice-based motion planning provides an opportunity to effective learning and monitoring of motion action execution due to the already discrete and well defined motion primitives (the actions). \emph{In this paper we present a general approach for enhancing lattice-based motion planning methods with (1) learning models of normal motion primitive execution, (2) using the learned models to improve collision checking effectiveness during planning and to (3) efficiently monitor the motion primitive execution for abnormalities.} The approach makes few assumptions and acts as a flexible plug-in to any modern AI-robotic software stack (Figure \ref{fig:schematic-overview}).
	
	Since both collision checking and abnormality detection are safety-critical the learning is performed using machine learning techniques with explicit uncertainty quantification from Bayesian machine learning \cite{lederer2019uniform}.
	%Since both collision checking and abnormality detection are safety-critical the learning is performed through machine learning techniques from Bayesian machine learning \cite{lederer2019uniform}.
	The monitoring for abnormalities also verifies at runtime that the learned models remain valid for collision checking use in the motion planner. %\TODO{The monitoring functionality for abnormality detection is presented as a ProbSTL plugin \cite{TIGER2020325}, making it ready for runtime verification using the proven correct algorithms of the ProbSTL framework.}
	
	%\TODO{Maybe mention more clearly that it is motion primitive execution from the robot's perspective. We monitor perception$\leftrightarrow$action integrity (plus), but can miss un-observable abnormal executions (minus).}
	%
	%\TODO{Safety margin. We are filling a hole (especially w.r.t. lattice planning). Place in subsection 'related work'?}
	%\TODO{Missing comparison/discussion w.r.t. other related work, e.g. \cite{park2016multimodal} and also w.r.t. for example adaptive control?}
	
	%Section \ref{ch:problem-formulation} describes modern lattice-based motion planning and control, and introduces the problem to be solved. Our approach for learning, improving collision checking with and monitoring motion primitive executions is presented in Section \ref{ch:our-approach}. Section \ref{ch:results} presents our results. Finally the conclusions and future work is presented in Section \ref{ch:conclusions}.
	Modern lattice-based motion planning and control given dynamic obstacles is described in Section \ref{ch:problem-formulation}. 
	Related work in \ref{ch:related-works}. Our approach for learning, improving collision checking and monitoring motion primitive executions is presented in Section \ref{ch:our-approach}, with results in \ref{ch:results} and conclusions in Section \ref{ch:conclusions}.

	\section{PROBLEM FORMULATION}
	\label{ch:problem-formulation}
	%\subsection{Execution Monitoring}
	%TODO: Highlight these aspects somewhere?
	%\TODO{Task planning and execution monitoring \cite{Doherty249047}. MTL specifications can be checked with runtime verification. Action: move-to(x). We can monitor many things: (1) If we reached x and/or within the time budget (completion condition), (2) collisions free execution and "normal" action execution (duration conditions). Hard to specify what normal is and too large safety-margins for the collision checking is detrimental to action effectiveness. Very difficult at the "action level" when actions are complex continuous motions.
	%\cite{park2016multimodal} does it somewhat but still very complex actions and many difficulties. We focus on the specific problem of motion control of a robot and uses the motion plan directly to reduce the complexity and to find efficient and effective methods for this problem using both learning and reasoning. (Move the latter to the Introduction.)
	%}

	\subsection{Motion Planning}
	\label{ch:problem-motion-planning}
	Consider a robot that is modeled as a time-invariant nonlinear system 
	\begin{align}
	\dot x(t) = f(x(t),u(t)),
	\label{eq:nonlinear_system}
	\end{align} 
	where $x(t)\in \mathbb R^n$ denotes the robot's states\footnote{For example position, velocity, orientation and angular velocity in 3D.} and $u(t)\in\mathbb R^m$ its control signals.
	The robot has physically imposed constraints on its states $x(t)\in\mathcal X$ and its control signals $u(t)\in\mathcal U$.
	The robot operates in a 2D or 3D world $\mathcal{W}(t)$, i.e. $\mathcal{W}(t)\subset \mathbb{R}^2$ or $\mathcal{W}(t)\subset \mathbb{R}^3$. There are regions $\mathcal{O}_{\text{obs}}(t)~\subset~\mathcal{W}(t)$ which are occupied by static and dynamic obstacles. Free space $\mathcal{X}_{\text{free}}(t) = \{x(t)\in\mathcal{X}\;\vert\;\mathcal{O}_{x(t)}\cap\mathcal{O}_{\text{obs}}(t) = \emptyset\}$ is where the region occupied by the robot $\mathcal{O}_{x(t)}$, transformed by the state $x(t)$, is not in collision with any obstacle at time $t$.
	
	%\TODO{
	%\textbf{Alternatives}
	%$$\mathcal{O}_{}\big(x(t)\big)$$
	%$$\mathcal{O}_{robot}\big(x(t)\big)$$
	%$$\mathcal{A}_{}\big(x(t)\big)$$
	%$$\mathcal{V}_{}\big(x(t)\big)$$
	%$$\mathcal{O}_{}(x_t)$$
	%$$\mathcal{O}_{robot}(x_t)$$
	%$$\mathcal{A}_{}(x_t)$$
	%$$\mathcal{V}_{}(x_t)$$
	%$$\mathcal{V}_{x}(t)$$
	%}
	%
	The objective of the motion planner is to produce a feasible reference trajectory $\big(x_0(t),u_0(t)\big)$, $t\in[t_S,t_G]$ that moves the robot from a starting state $x_S$ to a goal state $x_G$ while optimizing a given performance measure $J$, for example a compromise between minimum time and smoothness as in \cite{RHL-MP2018}. Taking the obstacles into account, the reference trajectory also must be collision free. This problem is called the dynamic motion planning problem (DMPP) \cite{RHL-MP2018}
	\begin{align}
	\minimize_{u_0(\cdot),\hspace{0.5ex}t_G}\hspace{1ex}
	& J = \int_{t_S}^{t_G}L(x_0(t),u_0(t),t)\,dt  \label{eq:MotionPlanningOCP}\\
	\subjectto\hspace{1ex}
	& \dot{x}_0(t) = f(x_0(t),u_0(t)),\nonumber  \\
	& x_0(t_S) = x_S, \quad x_0(t_G) = x_G, \nonumber \\ 
	& x_0(t) \in \mathcal X_{\text{free}}(t), \hspace{1ex} \forall t\in[t_S,t_G],\nonumber \\
	& u_0(t) \in \mathcal U, \hspace{1ex} \forall t\in[t_S,t_G], \nonumber
	\end{align}
	and it is typically intractable to even find a feasible solution.
	%\TODO{Say something about $L$. I.e. what it can be.}
	
	\subsection{Lattice-based Motion Planning}
	\label{ch:problem-lattice-motion-planning}
\begin{figure}[t]
		\vspace*{0.1in}
		\centering
		\includegraphics[width=0.98\columnwidth]{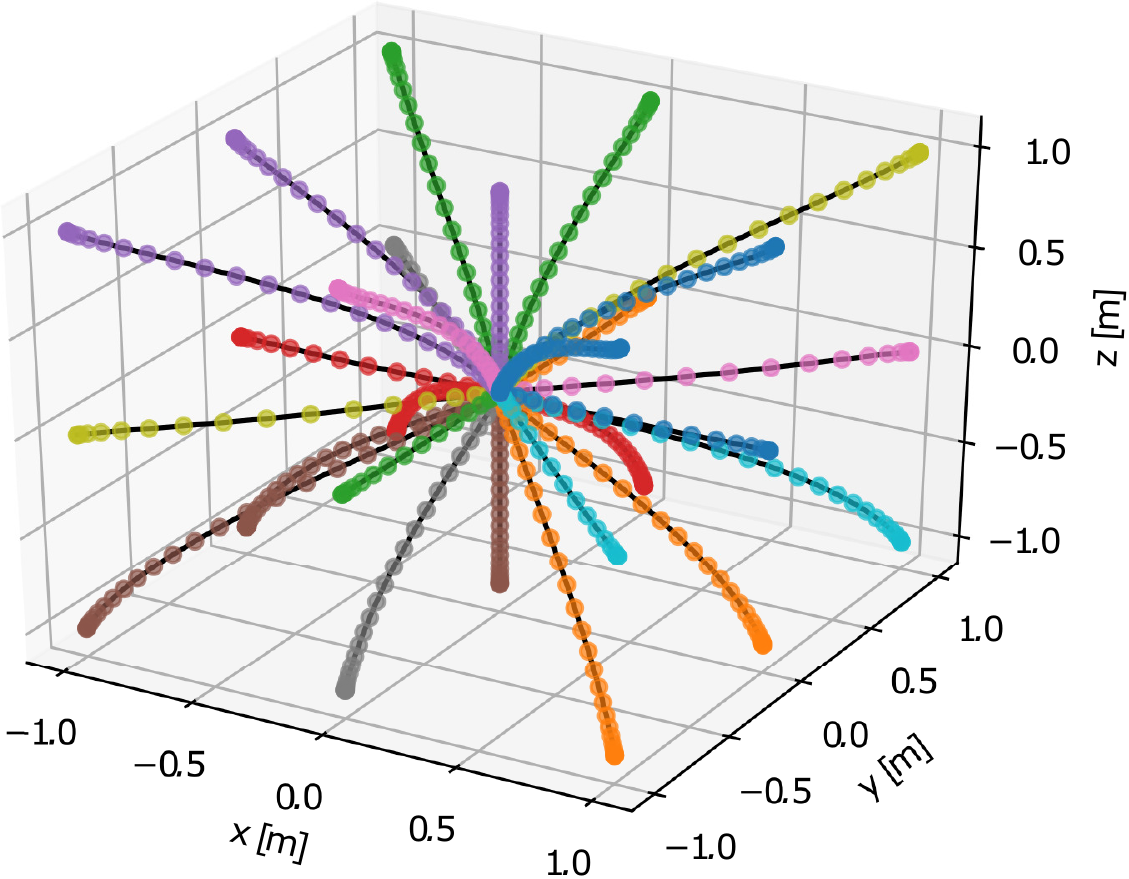}
		\caption{The 26 motion primitives with zero initial and final velocity, out of a total of 104 primitives. All start at position $[0,0,0]^T$.}
		%\caption{3D positions of 26 motion primitives, with initial and final velocity being zero. All start at position $[0,0,0]^T$. A total of 104 motion primitives are used in the experiments.}
		\label{fig:motion-primitives}
	\end{figure}

	% \cite{Katra2015} Describe Lattice-based motion planning (the basic)
	Lattice-based motion planning \cite{Katra2015} is a tractable approximation to DMPP \cite{RHL-MP2018} where the state space is discretized into a state lattice and a finite number of translation-invariant motion primitives (actions) are constructed to allow motion between states (nodes) on the lattice. Graph search techniques such as A$^*$ with an admissible heuristic can be used to find a valid sequence of motion primitive actions from $x_S$ to $x_G$. A motion primitive action $a_i\in\mathcal{A}$ is a trajectory $\big(x_0^i(t),u_0^i(t)\big),\;t\in[0,\,t_F^i]$ with initial state $x_I^i$ and final state $x_F^i$ on the lattice grid $\mathcal{X}_d \subset \mathcal{X}$ which satisfies the following
	\begin{subequations}
		\label{eq:primitive}
		\begin{align}
		\dot x_0^i(t) &= f(x_0^i(t),u_0^i(t)) \\
		x_0^i(0) &=x_I^i\in \mathcal X_d, \quad  x_0^i(t_F^i)=x_F^i\in \mathcal X_d \\
		x_0^i(t) &\in \mathcal X, \quad u_0^i(t) \in \mathcal U, \quad \forall t\in[0,t_F^i]
		\end{align}  
	\end{subequations}
	The motion primitives are generated offline leveraging numerical optimal control using the same loss function $L(x_0(t),u_0(t),t)$ and are assigned the resulting objective function value $J_i$. Figure \ref{fig:motion-primitives} show some motion primitives.
	
	The Lattice approximation to the DMPP is
	\begin{align}
	\minimize_{a_1,\dots,a_N,\hspace{0.5ex}N}\hspace{1ex}
	& J = \sum_{n=1}^N J_n	\label{eq:MotionPlanningLattice} \\
	\subjectto\hspace{1ex} 
	& \big(x_0^n(t),u_0^n(t)\big) = a_n \in \mathcal A , \quad \forall n,  \nonumber\\
	%& x_0^n(0) = x_0^{n-1}(t_F^n), \quad \forall n > 0, \nonumber \\
	& \big|\big|\mathrm{\mathbf{N}}x_0^n(0)- \mathrm{\mathbf{N}}x_0^{n-1}(t_F^n)\big|\big|_2 = 0, \quad \forall n > 0, \nonumber \\
	%& x^0_0(0) = x_I, \quad x^N_0(t^N_F) = x_G, \nonumber \\ 
	%& ||x^0_0(0) - x_I||_2 = ||\mathcal{O}_{x^0_0(0)} - \mathcal{O}_{x_I}||_2, \nonumber \\ 
	& \mathrm{\mathbf{T}}(x_I)\;x^0_0(0) = x_I, \nonumber \\ 
	%& ||x^N_0(t^N_F) - x_G||_2 = ||\mathcal{O}_{x^0_0(0)} - \mathcal{O}_{x_G}||_2, \nonumber \\ 
	& \mathrm{\mathbf{T}}_{N-1}\;x^N_0(t^N_F) = x_G, \nonumber \\ 
	& \mathrm{\mathbf{T}}_{n-1}\;x^n_0(t) \in \mathcal X_{\text{free}}(\mathcal{T}_{n-1}+t),\; \forall t\in[0,t^n_F], \; \forall n, \nonumber
	\end{align}
	where $\mathrm{\mathbf{T}}(x)$ is a translation matrix defined by the position in state $x$ and $\mathrm{\mathbf{N}}$ is a diagonal matrix with zeros at the position dimensions (projecting a state's position to zero under multiplication). $\mathrm{\mathbf{T}}_K = \mathrm{\mathbf{T}}(x_I)\prod_{k=0}^{K}\mathrm{\mathbf{T}}\big(x_0^k(t_F^k)\big)$ is the resulting translation of the first $K$ motion primitives in the plan and $\mathcal{T}_K = t_S + \sum_{k=0}^K t^k_F$ is the start time of the $K$:th motion primitive in the plan.
	
	The resulting plan $(t_S,x_I, a_0,\dots,a_N)$ consists of a sequence of $N$ motion primitive actions, $a_0,\dots,a_N,\;\forall a_n\in\mathcal{A}$. The end time of the plan is $t_G = \mathcal{T}_N$. The reference trajectory $(x_0(t),u_0(t)), t\in [t_S,t_G]$ is constructed from the plan by sequential spatio-temporal concatenation of the sequence of motion primitives in the plan.
	%
	% is the temporal concatenation of $u_0^n(t), \forall t\in[0,t^n_F],\;n = 1,\dots,N$ etc.
	%\TODO{Temporal concatenation of state and control reference trajectory.}
	%\begin{align} 
	%\minimize_{u_0[\cdot],\hspace{0.5ex}N}\hspace{1ex}
	%& \sum_{n=0}^{N}L(x_0[t_n],u_0[t_n],t_n) 	\nonumber \\
	%\subjectto\hspace{1ex}
	%& x_0(t_{n+1}) = f(x_0[t_n],u_0[t_n]), \label{MotionPlanningLatticeSearch} \\
	%& x_0(t_0) = x_I, \quad x_0(t_N) = x_G, \nonumber \\ 
	%& x_0[t_n] \in \mathcal X_{\text{free}}(t), \hspace{1ex} \forall n\in[0,N]\nonumber \\
	%& u_0[t_n] \in \mathcal U, \hspace{1ex} \forall n\in[0,N] \nonumber
	%\end{align}
	Lattice-based motion planning is resolution complete and is equivalent with the DMPP in the resolution-limit.
	
	The reference trajectory found by the motion planner is executed by a trajectory tracking controller, for example using a nonlinear Model Predictive Controller (MPC) \cite{RHL-MP2018}. The objective of the trajectory tracking controller is to 
	have the robot 
	follow the desired reference trajectory with a small tracking error $\tilde{x}(t) = x(t) - x_0(t)$ while keeping close to the feed-forward control signal $\tilde{u} = u(t) - u_0(t)$. The continuous-time nonlinear MPC problem, which is solved with respect to the current time point $t_0$, is formulated as \cite{RHL-MP2018}
	\begin{align}
	\minimize_{ u(\cdot)}\hspace{2.2ex} & ||\tilde{x}(t_0+T)||^2_{\mathbf P_N} + 
	\int_{t_0}^{t_0+T}\hspace{-1.5em}||\tilde{x}(t)||^2_{\mathbf R_1}  + ||\tilde{u}(t)||^2_{\mathbf R_2}\,dt \nonumber	\\
	\subjectto\hspace{1.2ex}
	& \dot{x}(t) = f(x(t),u(t)),  \label{OCPtraj}\\
	& u(t) \in \mathcal U, \nonumber
	\end{align}
	where the design parameters $\mathbf P_N, \mathbf R_1, \mathbf R_2$ are positive-definite weight matrices and $T$ is the prediction horizon.
	
	\subsection{Collision Checking}
	%TODO: Mention "Motion planning under uncertainty"
	The robot occupy a region $\mathcal{O}_{x(t)}$ that depends on its state (e.g. its position and orientation) at time point $t$. We want to find a reference trajectory $x_0(t)$ which at any time-point $t\in[t_S,t_G]$ is collision free $\mathcal{O}_{x_0(t)} \cap \mathcal{O}_{\text{obs}}(t) = \emptyset$.
	
	There exists many possible sources of uncertainty in applied motion planning. These can be due to e.g. modeling error, sensor noise, noisy control or the unpredictability of other agents. The uncertainty can be categorized into sensing of state and predictability of future state, due to for example robot actions, of both the robot and its environment \cite{lavalle1997motion}. A common way to alleviate the uncertainty is to use a safety margin to mitigate uncertainty induced collisions \cite{donald1993kinodynamic}.
	The kind of safety margin $\mathcal{O}_{\text{safety}}(t)\subset \mathcal{W}(t)$ considered in this work is a region relative to the robot's coordinate frame that extends the spatial occupancy of the robot, $$\mathcal{O}_{x_0(t)} = \mathcal{O}_{x_0(t)}^{*} * \mathcal{O}_{\text{safety}}(t),$$
	where $\mathcal{O}_{x_0(t)}^{*}$ is the actual occupied region of the robot and the binary operator $*$ over regions is defined as
	$$ \mathcal{O}_A * \mathcal{O}_B = \{p_A+p_B\,\vert\,p_A\in\mathcal{O}_A,\,p_B\in\mathcal{O}_B\}.$$
	Another approach is to extend the obstacle regions \cite{bergman2018combining}.
	
	The safety margin can be divided into three main parts,
	$$\mathcal{O}_{\text{safety}}(t) = \mathcal{O}_{\text{perception}}(t) * \mathcal{O}_{\text{control}}(t) * \mathcal{O}_{\text{others}}(t),$$
	reflecting the uncertainties from the robot's perspective of the state of the world, the control result and in the behavior of others. 
	In safety-critical applications where uncertainties are present, it is necessary to quantify the uncertainties explicitly as probability distributions in order to guarantee probabilistic safety \cite{lederer2019uniform} and probabilistic feasibility \cite{luders2010chance}. A safety margin can be constructed from such a representation to bound the probability of collision, e.g. to at most 0.01\% per second, frequently called \emph{chance constraint} in the 
	%\emph{motion planning under uncertainty} 
	literature \cite{luders2010chance}.
		
	%The safety-margin (e.g. radius) is 
	%%common
	%often left as a difficult design (or tuning) parameter in the literature. In some cases there are however systematic ways to decide the safety margin.	
	
	The third safety margin component $\mathcal{O}_{\text{control}}$ is important to lattice-based approaches but has thus far been largely overlooked.
	Due to modeling errors, noise sources and hardware limitations it is in practice unreasonable to expect the trajectory tracking controller will follow the reference trajectory perfectly. Even using exact models and having no noise, the accuracy depends on the motion primitive sequence, together with the choice of the design parameters $T, \mathbf P_N, \mathbf R_1, \mathbf R_2$. Such tracking errors tend to have both a bias and a covariance, and directly affect $\mathcal{O}_{\text{control}}$.
	In this paper we present a method for systematically deciding $\mathcal{O}_{\text{control}}$ for lattice-based motion planning, and also to runtime verify that $\mathcal{O}_{\text{control}}$ is indeed correctly probabilistically grounded. The latter to make sure that the safety guarantees are maintained in changing situations.

	\section{Related Work}
	\label{ch:related-works}
	The problem of real-time motion planning in cluttered, complex and dynamic environments has seen an increasing attention in the literature \cite{Katra2015}. Advances such as multi-resolution lattices \cite{pivtoraiko2009differentially} have made lattice-based approaches suitable for cluttered and complex environments. Lattice-based motion planning achieve state-of-the-art (SOTA) performance in for example trucks with multiple trailers \cite{Ljungqvist2019fieldrobotics}, and for quadcopters \cite{RHL-MP2018} for planning and re-planning, with both dynamics and time, in real-time with dynamic obstacles in complex environments.
		
	We outline the general problem setting of \cite{RHL-MP2018} in section \ref{ch:problem-motion-planning} and \ref{ch:problem-lattice-motion-planning}. We show how to integrate our proposed introspective capabilities to lattice-based motion planning by providing a detailed context, and our work is fully compatible with the multi-resolution lattice approach in \cite{RHL-MP2018}. The multi-resolution lattice approach in \cite{RHL-MP2018} is in turn based on \cite{pivtoraiko2009differentially} and the soft constraints in  \cite{pivtoraiko2009differentially} can be included in the loss function $J_n$.
		
	Common practice is to relax the composition of $\mathcal{O}_{\text{safety}}(t)$ and to pick an arbitrary large region as $\mathcal{O}_{\text{safety}}(t)$ likely to satisfy a desired safety constraint. The most common simplification of this kind is to use a spherical safety margin like \cite{CollisionAvoidETH17},  defined by a single radius parameter. This practice, although computationally efficient, generally require a much larger $\mathcal{O}_{x_0(t)}$ than what is actually necessary and consequently reduce collision checking effectiveness.
	This can be caused by different position uncertainty in different directions and position bias, both of which may change over time in strength and direction (e.g. strong wind).
	Previous work has investigated different parts of $\mathcal{O}_{\text{safety}}(t)$ and provided methodology to probabilistically ground both $\mathcal{O}_{\text{perception}}(t)$ and $\mathcal{O}_{\text{others}}(t)$.
	
	In \cite{gonzalez2018motion}, using lattice-based \emph{motion planning under uncertainty}, the state uncertainty is taken into consideration explicitly in the reference trajectory. A consequence is that $\mathcal{O}_{\text{perception}}(t)$ can be probabilistically grounded for every time point $t$ during planning. It can for example be as a $99\%$ probability region of the state for time point $t$. On its own this corresponds to a safety margin for which it is at least $99\%$ probable that the robot occupied region is within $\mathcal{O}_{x_0(t)}$.
	
	If a realistic simulator of the target domain is available it can be used to simulate realistic behaviors of other agents, and the interaction between such agents and the robot. Such a simulator can be used to ground the uncertainty of $\mathcal{O}_{\text{others}}(t)$ and the margin can be tuned to satisfy for example a $99.9\%$ probability of no collision \cite{andersson2016model}. Since $\mathcal{O}_{\text{others}}(t)$ is grounded in predictive models of other agents, it is applicable to various motion planning approaches including lattice-based methods.
	
	The safety margin part $\mathcal{O}_{\text{control}}$ has been considered in other motion planning approaches, such as those based on Rapidly-expanding Random Trees (RRT), where process noise is used to sample possible motion plan executions \cite{luders2010chance}. In \cite{gonzalez2018motion} the process noise affects the covariance around the reference trajectory, but the uncertainty is unbiased.

	Our contributions, apart from the presented integration into lattice-based motion planning, are useful outside of the lattice-based paradigm. Learning models and abnormality detection of motion primitives assume that there is a fixed set of motion primitives and that we can observe their execution, such as in SOTA for quadcopters motion planning \cite{zhou2019robust,ryll2019efficient}. The improved effectiveness of collision checking further assumes that the models are a precise description of execution variability, which may not always hold with post-search trajectory optimization \cite{Oliveira2018,zhou2019robust}. This would cause our proposed method to also include the variation in different optimization results.

	\section{INTROSPECTIVE LATTICE-BASED MOTION PLANNING AND CONTROL}
	\label{ch:our-approach}
	
	\begin{figure}[h!]
		\includegraphics[width=\columnwidth]{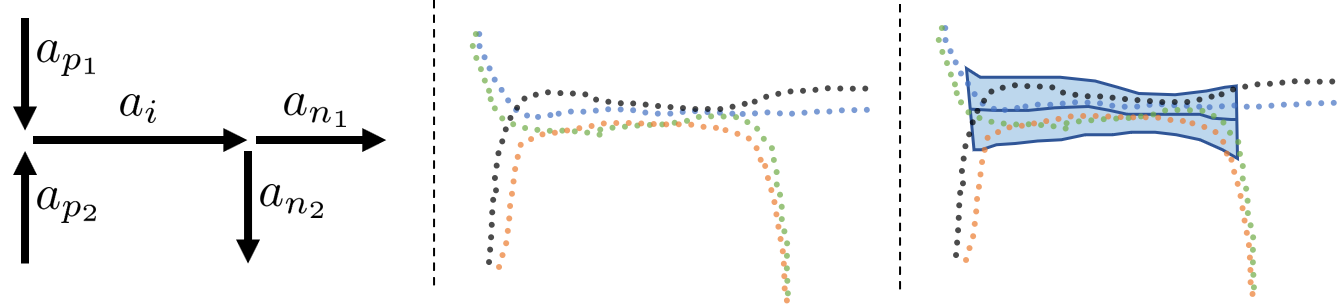}
		\caption{Illustration of learning a model of the \emph{normal} execution of primitive $a_i$. \textbf{Left:} All valid plans of triplets with $a_i$ in the middle ($a_{p_*}$ previous and $a_{n_*}$ next). \textbf{Center:} Observed execution of each triplets plan. \textbf{Right:} Mean predictive and 95\% probability region of unimodal distribution (\ref{eq:HPGP-predictive}) capturing the expected variability of normal $a_i$ executions.}
		\label{fig:model-learning-illustration}
	\end{figure}

	\subsection{Learning Normal Primitive Execution}	
	The execution of a motion primitive $a_i\in\mathcal{A}$, as perceived by the robot, is a discrete trajectory $\trajx = \hat{x}_{t_0},\hat{x}_{t_1},\dots$ with time points $\trajt = t_0,t_1,\dots$.
	Let $a_p\in\mathcal{A}$ and $a_n\in\mathcal{A}$ denote a valid previous and next motion primitive with respect to $a_i$ such that $[a_p,a_i,a_n]$ is a valid plan. For every $a_i\in\mathcal{A}$, let $\trajX^i = \trajx^1,\dots,\trajx^{J_i}$
	% and $\trajT_i = \trajt_1,\dots,\trajt_{J_i}$
	denote the $J_i$ observed executions of primitive $a_i$, one for every valid triplet $[a_p,a_i,a_n]$. All executions of primitive $a_i$ have the same time duration. Note that the time-points will not be aligned, and the number of data-points might vary slightly between executions. An illustration of the model learning steps are shown in Figure \ref{fig:model-learning-illustration}.

	To recover $x(t)$ for the execution of primitive $a_i$ from $\trajx^j$ we assume a nonlinear additive Gaussian noise regression model with diagonal covariance $\Sigma=\text{diagonal}\big(\sigma^2_{0},\,\sigma^2_{1},\dots\big)$,
	\begin{equation*}
	\hat{x} = x(t) + \epsilon, \quad\quad \epsilon \sim \mathcal{N}(0, \Sigma),
	\end{equation*}
	and place a Gaussian process prior on the function $x$
	\begin{equation*}
	x \sim \mathcal{GP}(m(\cdot), k(\cdot,\cdot)).
	\end{equation*}
	A Gaussian process \cite{Rasmussen06gaussianprocesses} is a distribution over functions which has been highly successful in many statistical analysis and regression tasks, for example for motion pattern recognition \cite{Kim2011}.
	%TODO: Remove reference ^?
	It is a Bayesian non-parametric model which is very suitable for modeling trajectories and trajectory-based motion patterns \cite{TigerFUSION2015}. The Gaussian process is defined by mean function $m(t)$ and covariance function $k(t_1,t_2)$. 
	By conditioning the GP on the observed trajectory $\trajx^j$ we get a predictive distribution where, for every time point $t$,
	\begin{equation}
	p(x\vert t, \trajx^j) = \mathcal{N}\big(x\,\vert\, \mu_{\trajx^j}(t),\,\Sigma_{\trajx^j}(t)\big),
	\label{eq:GP-predictive}
	\end{equation}
	where, over all state dimensions $d = 1,\dots,n$,
	\begin{align}
	\mu_{\trajx^j}(t) &= [\dots,\;\mu_{\trajx^j_d}(t),\dots]^T,\\
	\Sigma_{\trajx^j}(t) &= \text{diagonal}\big(\dots,\,\sigma^2_{\trajx^j_d}(t),\dots\big),
	\end{align}
	where, using a zero mean function $m(t) = 0$ since we can readily subtract the mean from the data,
	\begin{align}
	\mu_{\trajx^j_d}(t) &= \textrm{K}({t},\trajt^j)^T\textrm{V}^{-1}_d(\trajx^j_d),
	\label{eq:GP_mean}\\
	\sigma^2_{\trajx^j_d}(t) &= \textrm{K}({t}, {t}) + \sigma^2_{d} - \textrm{K}({t},\trajt^j)^T \textrm{V}^{-1}_d \textrm{K}({t},\trajt^j),
	\label{eq:GP_covariance}
	\end{align}
	where $[\textrm{K}]_{ab} = k(t_a,t_b)$, $\textrm{V}_d = \textrm{K}(\trajt^j,\trajt^j) + \sigma^2_{d}\textrm{I}_{\trajt^j}$,
	$\textrm{I}_{\trajt^j}$ is an identity matrix. Since $t$ is a scalar then $\textrm{K}({t}, {t})$ is a scalar,  $\textrm{K}({t},\trajt^j)$ is a column vector and $\textrm{V}_d$ a square matrix. Each output dimension is for simplicity treated as being independent which is equivalent to them being modeled by a separate function each with a separate Gaussian process prior.
		
	Using (\ref{eq:GP-predictive}) it is now possible to align all $J_i$ triplet executions over the same time interval $[0,t_F^i]$. This set of executions spans the variety of primitive $a_i$ and we expect any future \emph{normal} execution of $a_i$ to be similar to this set.
	
	A motion primitive is a 1D curve through $n$D state space parameterized by time. We therefore consider a unimodal distribution over functions (like a single GP) a suitable model to represent the execution variability of $a_i$ \cite{TigerFUSION2015}. 
	
\begin{figure}[h!]
		\centering
		\includegraphics[width=\columnwidth]{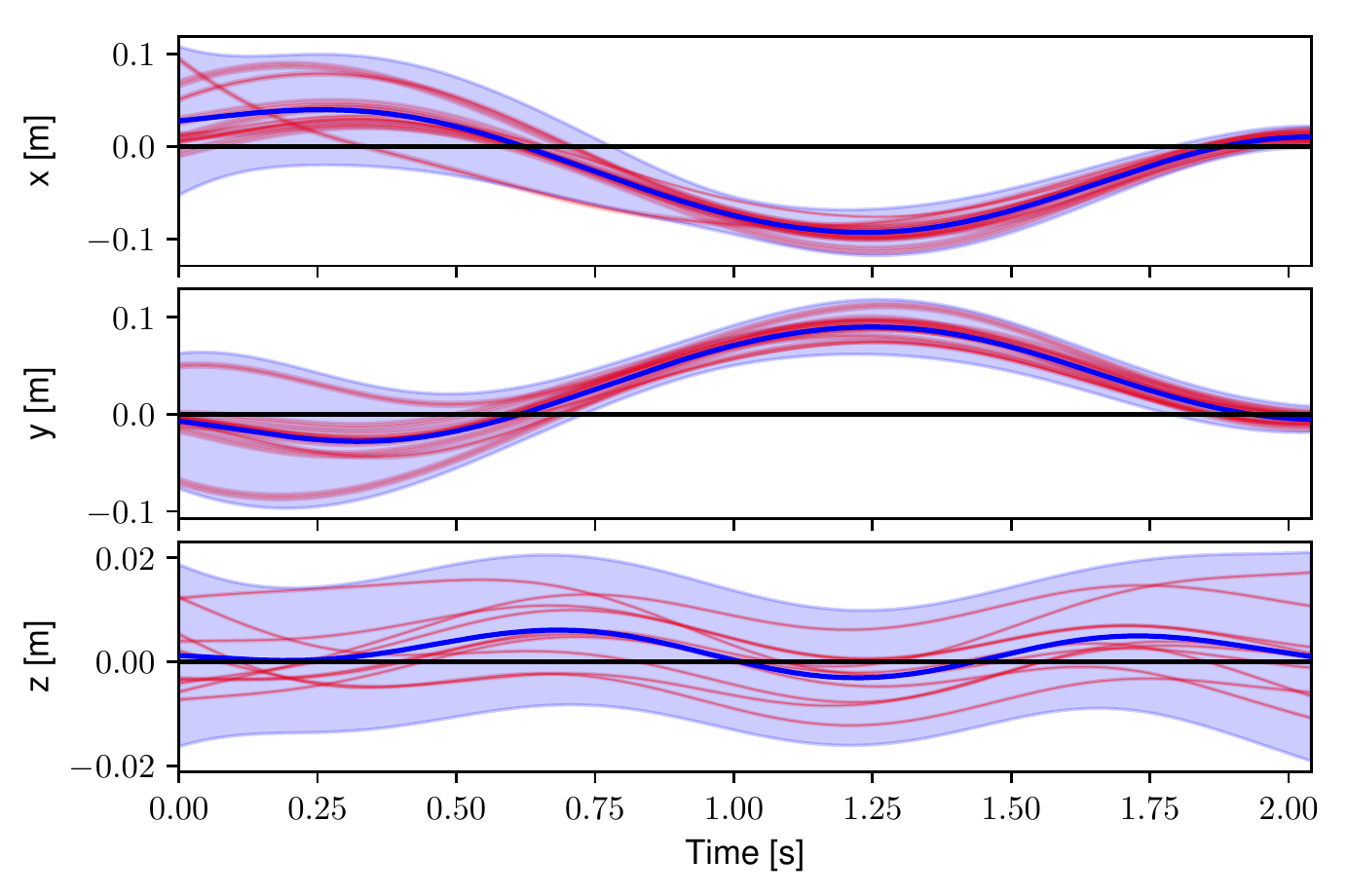}
		\caption{Learned unimodal distribution for a motion primitive. The plot shows residuals w.r.t. the reference state trajectory of the motion primitive, i.e. $x^i_0(t)$ is subtracted. The red lines are the individual Gaussian processes for the observations. The unimodal distribution is defined by the blue region, displaying its 99\% probability region, and the blue line is its mode.}
		\label{fig:model-learning-example}
	\end{figure}

	We formulate the \emph{motion primitive execution model} as a unimodal distribution over functions
	\begin{equation}
	p(x\vert t, \trajX^{i}) = \mathcal{N}\big(x\,\vert\, \mu_i(t),\,\Sigma_i(t)\big)
	\label{eq:HPGP-predictive}
	\end{equation}
	where $\mu_i(t)$ and $\Sigma_i(t)$ are given, per dimension, by \cite{TigerFUSION2015}
	%\\
	%\TODO{Add State estimation covariance matrix (or rather the marginalized variance in every dimension) to $\Sigma_i(t)$?}
	%\TODO{Say something about where we get the GP output data points (State Estimation / Bayesian filter point estimate)?}
	\begin{align}
	\mu_{i}(t) &= 
	\frac{1}{J_i}\sum_{m = 1}^{J_i}\mu_{\trajx^j}(t),
	\label{eq:gaussians_combined}\\
	\sigma^2_{i}(t) &= 
	\frac{1} {J_i}\sum_{m = 1}^{J_i}\big(\sigma^2_{\trajx^j}(t) + \mu_{\trajx^j}(t)\big) - \mu_{i}^2(t),\nonumber
	\end{align}
	which can be interpreted as the sample mean and sample variance from noisy samples with Gaussian-distributed noise.
	It is a \emph{Gaussian approximation} to a mixture of Gaussian Processes, MoGPs. The MoGPs are based on the set of GPs all with the same weight (every triplet is observed once). The Gaussian approximation allows the MoGP to generalize outside of the individual executions. The motion primitive execution model is illustrated in the third figure to the right in Figure \ref{fig:model-learning-illustration} and shown for 3D position in Figure \ref{fig:model-learning-example} together with the motion primitive state trajectory and observed executions.
	
	\subsection{Collision Checking}
	\label{ch:collision-checking-theory}
	The learned motion primitive models represents the state of the robot when executing the motion primitive. Its mode is equally or more accurate, on average, than the motion primitive reference state trajectory under the assumption of Gaussian process noise. The execution variability is explicitly represented as a probability density over the state, over time.
	
	For a stochastic variable $x$ distributed according to a multivariate Gaussian distribution, $x \sim \mathcal{N}(\mu,\,\Sigma)$, the centered \emph{probability region} (PR) with probability $\prob$ is defined as \cite{chew1966confidence}
	\begin{align}
	\mathrm{PR}(\prob,\,\mu,\,\Sigma) = \Big\{x \;\vert\; (x-\mu)^T\Sigma^{-1}(x-\mu) \leq \chi^{2}_{n}(\prob)\Big\},\nonumber
	\end{align}
	where $\chi^{2}_{n}(\prob)$ is the chi-square quantile function with $n$ degrees of freedom for probability $\prob$. For a diagonal $\Sigma = \text{diagonal}(\dots, \sigma^2_d, \dots)$ it simplifies to a axis-aligned ellipsoid,
	\begin{align}
	\mathrm{PR}(\prob,\,\mu,\,\sigma^2) = \Big\{x \;\vert\; \sum_d{(x_d-\mu_{d})^2}/{\sigma^2_d} \leq \chi^{2}_{n}(\prob)\Big\},\nonumber
	\end{align}
	with axis lengths given by $\sqrt{\sigma^2_d\chi^{2}_{n}(\prob)}$ for each axis $d$.
	
	Let $K$ be a motion primitive in a motion plan (\ref{eq:MotionPlanningLattice}), corresponding to motion primitive $a_i\in\mathcal{A}$ and with start time $\mathcal{T}_K$. Let $\tau = t-\mathcal{T}_K$ which limits $\tau$ between $0$ and $t^i_F$. The model $p(x\vert t, \trajX^{i})$ of motion primitive $a_i$ is a $n$-dimensional multivariate distribution with diagonal covariance matrix at each time-point $\tau$. The safety margin $\mathcal{O}_{\text{control}}$ can now be grounded probabilistically given probability $\prob$,
	\begin{align}
	&\mathcal{O}_{\text{control}}(\tau) = \mathrm{PR}\big(\prob,\,\mu_i(\tau),\,\sigma^2_i(\tau)\big)
	\end{align}
	
	\subsection{Abnormality Detection}
	Once we have learned the motion primitive models, we want to use them to to detect abnormal executions of motion primitives.
	A probability region over the execution variability model of a primitive describes in which states we expect the robot to be in for a given time point. For example \cite{TIGER2020325} uses probability interval 
	in execution monitoring for robot safety.

	There are several possible ways to use the probability region to define what normal behavior is.
	A naive approach is to check whether the robot ever leaves the region, and if it does, define that execution as abnormal.
	However, since we pick the region to not contain all probability, we still expect the robot to leave the probability region occasionally.
	
	More precisely, if we pick a 99\% PR, the robot is expected to leave the interval in 1\% of the time points.
	If the robot starts leaving the probability region more frequently, it might be an indication of an error.
	We define the rate of leaving the probability region as the failure rate, and model it as a stochastic variable.
	We consider the following parameters:
	\begingroup
	\allowdisplaybreaks
	\begin{align*}
	\theta  &-  \text{Failure rate}\\
	W &-  \text{Time window $t_W$ seconds into the past}\\
	\text{\#normal}_W &-  \text{The number of normal observations}\\
	\text{\#abnormal}_W &-  \text{The number of abnormal observations}\\
	\text{N}_W &-  \text{\#normal}_W + \text{\#abnormal}_W\\
	\alpha &-  \text{Prior belief about failure rate}\\
	\beta &-  \text{Prior belief about success rate}\\
	\end{align*}
	\endgroup
	
	We encode our prior assumptions about $\theta$ as a Beta distribution, setting the mode to $1 - \prob$, where e.g. $\prob=99\%$,  while still allowing for some uncertainty:
	$$\theta \sim \text{Beta}(\alpha,\beta),$$
	with $\alpha = N \prob$ and $\beta = N (1 - \prob)$, where $N$ is the prior strength.
	
	Given that we know $\theta$, the number of observed failures follows a Binomial distribution:
	$$\text{\#abnormal}_W\;\vert\;\text{N}_W,\theta \;\sim \text{Binomial}(\text{N}_W,\theta)$$
	
	As the Beta distribution is a conjugate prior for the Bionomial distribution, the posterior for $\theta$ is also Beta:
	\begin{align*}
	&\theta\;\vert\;\text{\#abnormal}_W,\text{\#normal}_W \\ 
	&\quad\quad\quad\quad \sim \text{Beta}(\alpha+\text{\#abnormal}_W,\beta+\text{\#normal}_W)
	\end{align*}
	
	Using the posterior for $\theta$, it is possible to evaluate the probability of $\theta$ being larger than expected:
	\begin{align*}&p(\theta > (1 - \prob)\;\vert\;\text{\#abnormal}_W,\text{\#normal}_W) \\
	& \; = 1 - \text{Beta}_{\text{CDF}}(1-\prob,\,\alpha + \text{\#abnormal}_W,\,\beta+\text{\#normal}_W)
	\end{align*}
	where $\text{Beta}_{\text{CDF}}(x,a,b)$ is a constant-time standard function in most statistics libraries,
	\begin{equation*}
	\text{Beta}_{\text{CDF}}(x,a,b) = \int_0^x q^{a-1}(1-q)^{b-1} dq.
	\end{equation*}
	
	This probability of the failure rate being higher than $(1~-~\prob)$ can then be monitored and thresholded.
	In this work we use the threshold of 99.9\% probability and the prior strength of $N = 1000$, but these values can be fine-tuned to select a trade-off between precision and recall.
	
	%\subsection{ProbSTL with discrete-continuous hybrid actions}
	%\TODO{remove}
	%ProbSTL \cite{TIGER2020325} extended with an action domain $\mathcal{A}$ consisting of the motion primitives, together with the \textit{normal} model of planned $a\in\mathcal{A}$, $a_{t\vert t}$, and the signal consisting of the perceived physical state of executing $a$, $a_t^{\text{observed}}$:
	%\begin{align}
	%	\Box\;\forall a\in\mathcal{A}\Big[\text{Pr}(\text{insidePI}(a_{t\vert t},a^{\text{observed}}_t,0.99)) \geq 0.99 \Big]
	%\end{align}
	%For all time points it holds that for every action $a$ (motion primitive) the probability of the execution of $a$ falling inside the 99\% probability region of the planned $a$ has to be at least 99\% probable.
	%
	%
	%\subsection{Approximation}
	%\TODO{Remove, or move to results?}
	%If discretization of GP-based primitive-execution model is used: Let MPC frequency be $F=50hz$, upsample-factor be $K=19$, number of actions $N=200$ and max-length in time of any action $T=3seconds$:
	%\begin{equation}
	%NTFK = 200*3*50*10 = 300000
	%\end{equation}
	%For dimensionality $D=6$ and both $\mu$ and $\sigma^2$ for every time point stored as $Float32$ the upper bound of the space requirement is
	%\begin{equation}
	%2DNTFK = D*600000\text{ bytes} = 3.6\text{Mb}
	%\end{equation}
	
	\section{EXPERIMENTS AND RESULTS}
	\label{ch:results}
\begin{figure}[h]
		\vspace*{0.1in}
		\includegraphics[width=\columnwidth]{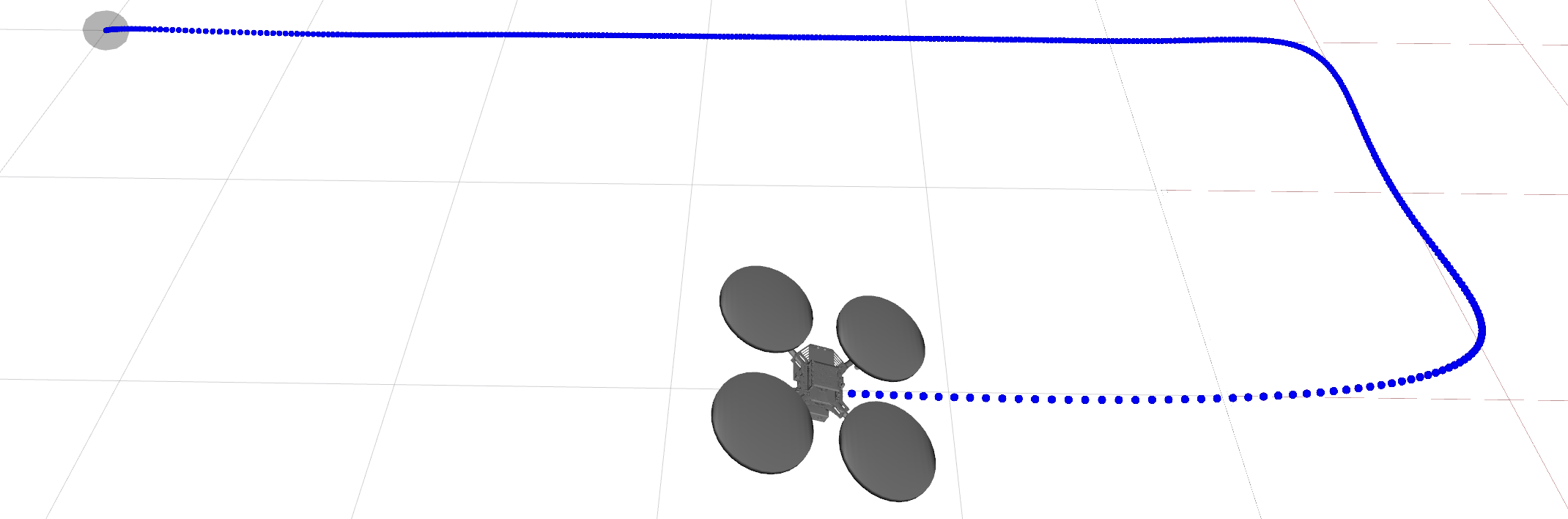}
		\caption{Simulated DJI Matrice 100 quadcopter used in this work.}
		\label{fig:simulated-M100}
	\end{figure}

	In order to evaluate our proposed approach we consider a simulated DJI Matrice 100 quadcopter (Figure \ref{fig:simulated-M100}), a commonly used commercial research platform.
	
	\subsection{Motion Planning and Control}
	The same non-linear model for the DJI Matrice 100 as in \cite{RHL-MP2018} is used. We use a nonlinear MPC controller based on the work in \cite{eth_mav_control} and use ACADO \cite{AcadoToolkit} to generate an efficient implementation that solves (\ref{OCPtraj}).
	A total of 104 motion primitives are generated using ACADO (Figure \ref{fig:motion-primitives}) using the same objective as the nonlinear MPC controller. 
	Let $\mathcal{Z} = \big\{(p,v)\,\vert\,v = p/||p||_2,\; p\in\{-1,0,1\}^3\backslash\{\textbf{0}\}\big\}$. The motion primitives have the initial and final states:
	\begin{itemize}
		\item From $x_I = (\textbf{0},\textbf{0})$ to $x_F = (p,\textbf{0})$, $\forall (p,v)\in\mathcal{Z}$
		\item From $x_I = (\textbf{0},\textbf{0})$ to $x_F = (p,v)$, $\forall (p,v)\in\mathcal{Z}$
		\item From $x_I = (\textbf{0},v)$ to $x_F = (p,\textbf{0})$, $\forall (p,v)\in\mathcal{Z}$
		\item From $x_I = (\textbf{0},v)$ to $x_F = (p,v)$, $\forall (p,v)\in\mathcal{Z}$
	\end{itemize}
	
	Running a sequence of triplet executions may cause one triplet execution to interfere with the next one. To avoid transient noise propagating across triplets, each triplet plan is augmented with actions before and after the triplet such that the full plan is always started from a state of rest.
	
	\subsection{Learning}
	For the motion primitive models, a Gaussian process prior with the squared exponential covariance function is used,
	\begin{equation}
	k(t_1,t_2) = \sigma^2_f e^{(-\frac{1}{2}(t_1-t_2)\Lambda(t_1-t_2)^T)},% + \sigma^2_n\delta(x_1,x_2)
	\label{eq:se-kernel}
	\end{equation}
	where $\Lambda$ is a diagonal matrix with length scales for each input dimension and $\sigma^2_f$ is the signal variance. These, together with the diagonal noise covariance $\Sigma$, are the hyper parameters $\theta~=~\{\sigma^2_f$, $\Sigma$, $\Lambda\}$ for this hierarchical Bayesian model.
	
	We estimate the hyper-parameters from the data by maximizing the marginal log likelihood (empirical Bayes),
	\begin{equation}
	\log p(x | t, \theta) = -\frac{1}{2}x^T\textrm{V}^{-1}x-\frac{1}{2}\log |\textrm{V}|
	+ C,
	\label{eq:log-likelihood}
	\end{equation}
	where $\text{V}$ is defined as in (\ref{eq:GP_mean}-\ref{eq:GP_covariance}) and $C$ is a constant.
	
	We investigate how close the predictive mean of the learned motion primitive execution model is to the observed trajectories, and compare it to how close motion primitive reference state trajectory is to the observations.
	In Table \ref{table:experiments-rmse} the first row shows the RMSE between observed executions and the reference state trajectory or the executed motion primitive. The mean predictive is significantly more accurate as a point prediction for the execution trajectory than the reference state trajectory, as can be seen in the row below. The reason for this is that the reference state trajectory has a larger bias to the mean over execution trajectories (across state, not time) for a single primitives in comparison with the learned model. An example can be seen in Figure \ref{fig:model-learning-example}.
	
	The mean predictive of the learned motion primitive model and the primitive's reference state trajectory are compared in the third row in Table \ref{table:experiments-rmse}. It is observed that the RMSE between them is almost the same as the RMSE between the reference state trajectory and the observed executions. This provide additional evidence that the learned motion primitive model is a systematically better mean predictor for the execution, compared to the motion primitive itself.
	
	\begin{table}[t]
		\vspace*{0.1in}
		\setlength\tabcolsep{0pt} % let LaTeX figure out amount of inter-column whitespace
		\caption{}
		\label{table:experiments-rmse}
		\sisetup{detect-weight=true,detect-inline-weight=math}
		\footnotesize 
		\begin{tabular*}{\columnwidth}{@{\extracolsep{\fill}} l r}
			\toprule
			\textit{} & \multicolumn{1}{c}{$\text{RMSE}$ (\textit{meters})}\\
			\midrule
			Observed Executions vs Motion Primitive			& 0.149 $\pm$ 0.053\\
			Observed Executions vs $\mu_i(t)$		 		& \textbf{0.055} $\pm$ 0.043\\\midrule
			Motion Primitive vs $\mu_i(t)$		            & 0.143 $\pm$ 0.052\\
			\midrule
		\end{tabular*}
		\caption*{The table shows several root-mean-square errors (RMSE). \textbf{Row 1:} The error between the motion primitive reference state trajectory and the observed executions of the same motion primitive. \textbf{Row 2:} The error between the mean prediction of the model of motion primitive execution and the observed executions of the same motion primitive. \textbf{Row~3:} The error between the primitive reference trajectory and the mean prediction of the model of executing that motion. \\Only positions from the state are considered here. The RMSE mean and standard deviation is over all motion primitives.}
	\end{table}

	\subsection{Collision Checking}
	The standard approach in the literature is to use a sphere for $\mathcal{O}_{\text{safety}}(t)$ \cite{CollisionAvoidETH17}, capturing the aggregation of $\mathcal{O}_{\text{perception}}(t)$, $\mathcal{O}_{\text{control}}(t)$, $\mathcal{O}_{\text{others}}(t)$, with a single radius parameter invariant to time.
	We use this baseline, $R$, and construct two more with tighter bounds, $R_i$ (different for each primitive) and $R_i(\tau)$ (also varying over time). For a meaningful comparison with our proposed variability model, we probabilistically ground the baselines by assuming a Gaussian likelihood over the observations centered on the primitive reference trajectory (i.e. relative translation is zero). For $R_*$ in $\{R,\;R_i,\;R_i(\tau)\}$,
	\begin{align}
	R_*\sim \mathcal{N}\big(0,\,\sigma^2_{R_*}\big),
	\end{align}
	where $\sigma^2_{R_*}$ for respective baselines are
	\begin{align}
	&\sigma^2_{R_i(\tau)} &&= \frac{1}{J_i-1}\sum^{J_i}_{j=1} \big(\mu_{\trajX^{i}}(\tau) - x^i_0(\tau)\big)^2\\
	&\sigma^2_{R_i} &&= \max_\tau\; \sigma^2_{R_i(\tau)}\\
	&\sigma^2_{R} &&= \max_i\; \sigma^2_{R_i}
	\end{align}
	separate for each dimension $d$, and $\tau$ as defined in \ref{ch:collision-checking-theory}.
	The safety margin $\mathcal{O}_{\text{control}}$, given probability $\prob$ is 
	\begin{align}
	&\mathcal{O}_{\text{control}}(\tau) = \mathrm{PR}(\prob,\,\textbf{0},\,\sigma^2_{R_*})
	\end{align}
	
	In Table \ref{table:experiments-area} the baselines are compared with the learned motion primitive model.
	$R$ is the largest 99\% probability region over all primitives, per dimension, centered at the reference trajectory. $R_i$ is specific to the individual primitives which allows it to be smaller. $R_i(t)$ is time variant, allowing a further radius reduction where the observations allows it.
	
\begin{figure}[b!]
		\centering
		\includegraphics[width=\columnwidth]{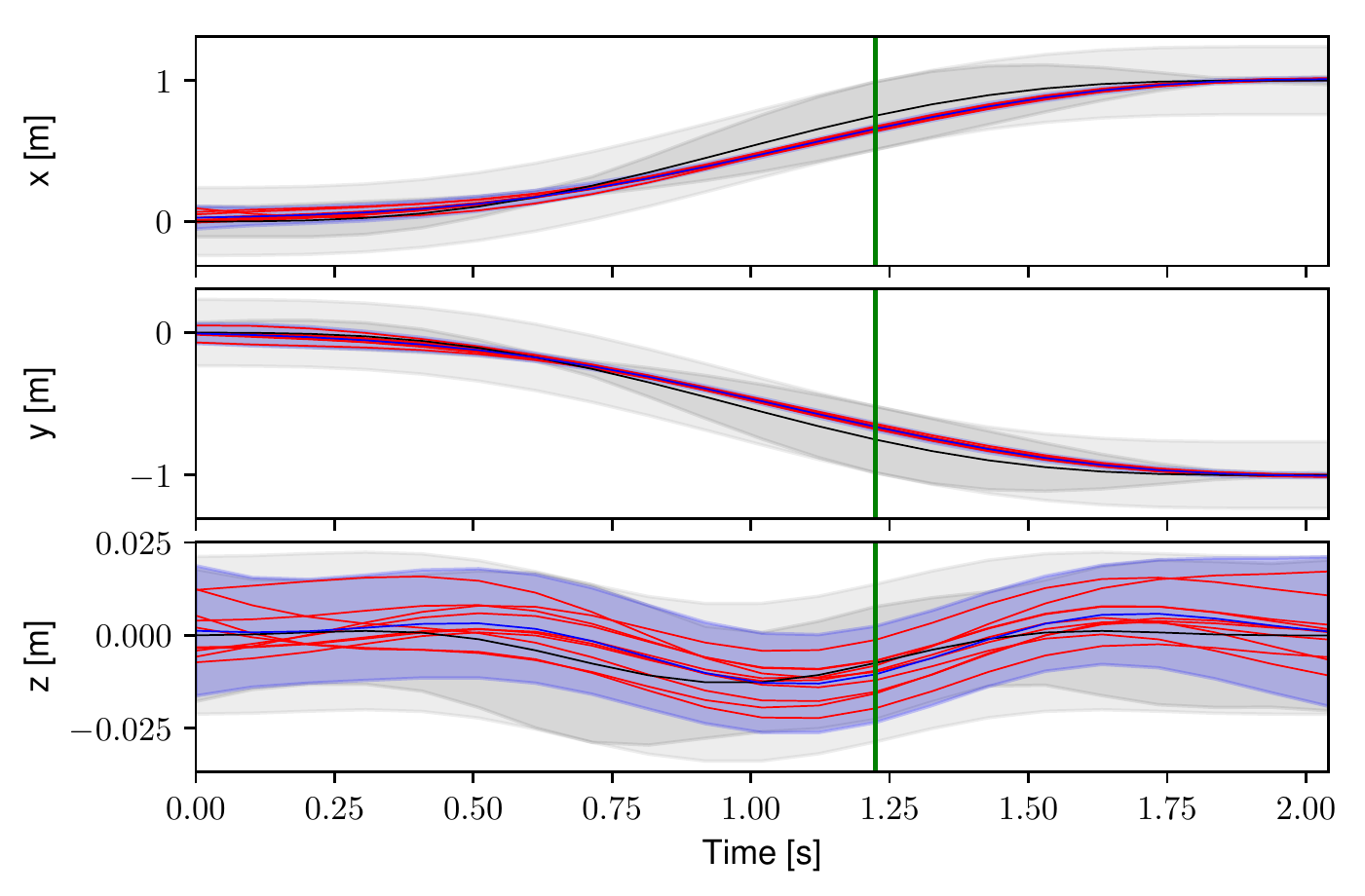}
		\caption{A comparison of $R_i$ (light gray envelope), $R_i(t)$ (dark gray
			envelope), $p(x\vert t, \trajX^{i})$ (blue line and blue
			envelope) and observed executions (red) of a single motion primitive $i$ (c.f. Fig. \ref{fig:model-learning-example}).
			$R_i$ and $R_i(t)$ share the same black line as their mean (the reference
			state trajectory). $R$ is not shown in order to make the finer details of the others visible. Envelopes are $2.6\sigma$.}
		\label{fig:experiment-example-area}
	\end{figure}

\begin{figure}[t!]
		\vspace*{0.1in}
		\centering
		\includegraphics[width=\columnwidth]{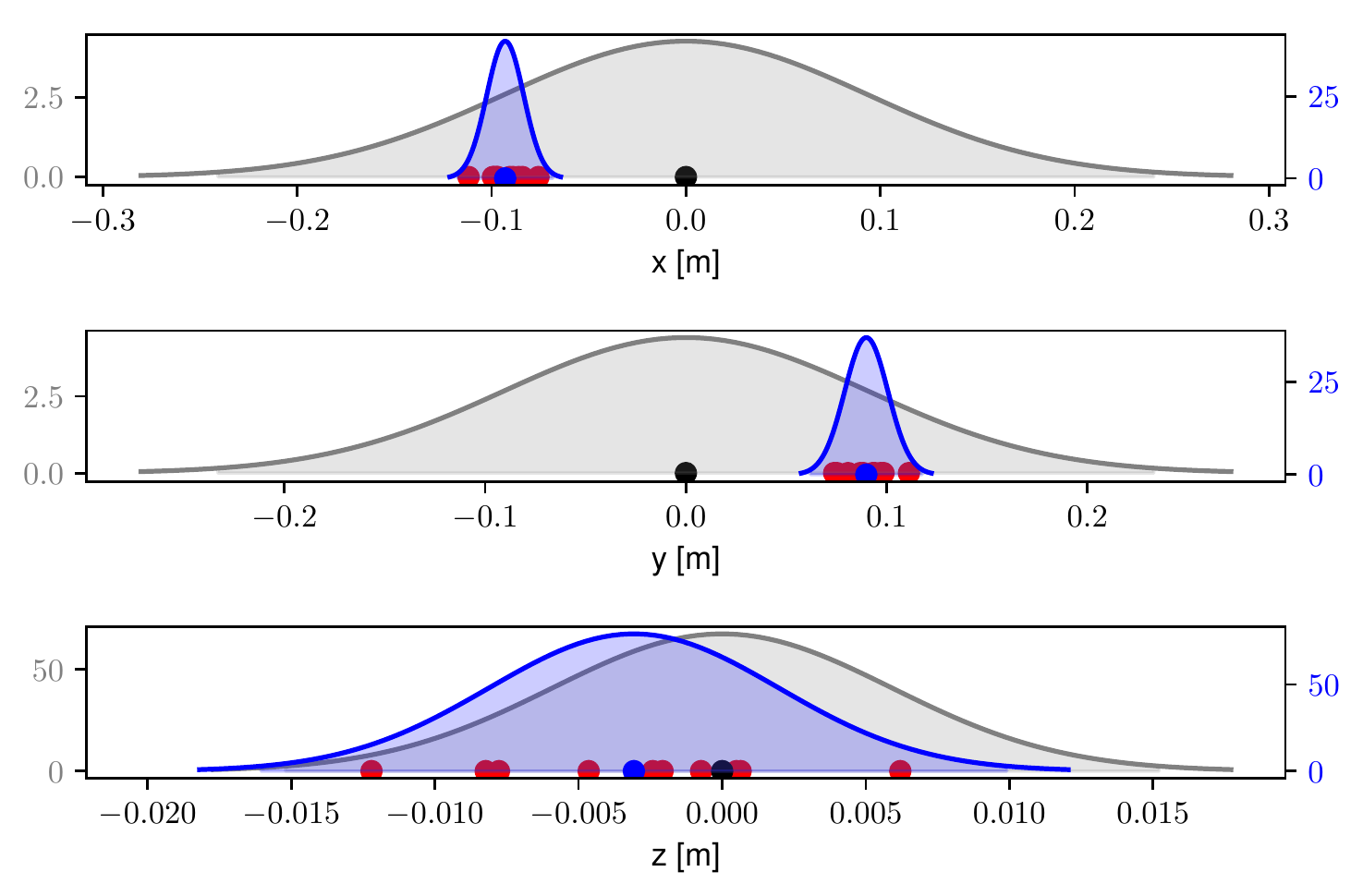}
		\caption{Residual plots of dissection of Figure \ref{fig:experiment-example-area} at $t=1.2$ (indicated with a vertical green line). $R_i(t)$ (black dot and dark gray envelope) and $p(x\vert t, \trajX^{i})$ (blue dot and blue envelope) are depicted together with observed executions (red dots) of a single motion primitive $i$. $R$ and $R_i$ are not shown as to make the finer details of the others visible. Envelopes are $2.6\sigma$.}
		\label{fig:experiment-example-area-disection}
	\end{figure}

	Figure \ref{fig:experiment-example-area} show an example comparing $R_i$, $R_i(t)$ and the learned model $p(x\vert t, \trajX^{i})$ for one motion primitive.
	Figure \ref{fig:experiment-example-area-disection} show the dissection of Figure \ref{fig:experiment-example-area} at the green vertical line.
	A symmetric safety margin around a center that has a large bias with respect to the mean can be detrimental. To still ensure safety, e.g. that observed executions should still be within $\prob~=~99\%$, the margin must be larger with larger bias. 
	
	A symmetric safety margin is at its smallest when centered at the mean. 	
	Table \ref{table:experiments-area} show that the area reduction is large using the learned variability model compared to the baseline. Even when comparing with a time-varying radius, $R_i(t)$, the area is reduced more than 3 times when using the learned model.
	
\begin{table}[t]
	\vspace*{0.1in}
	\setlength\tabcolsep{0pt} % let LaTeX figure out amount of inter-column whitespace
	\caption{}
	\label{table:experiments-area}
	\sisetup{detect-weight=true,detect-inline-weight=math}
	\footnotesize 
	\begin{tabular*}{\columnwidth}{@{\extracolsep{\fill}} l l r}
		\toprule
		% & \multicolumn{3}{c}{Metric}\\
		\textit{} &  \multicolumn{2}{r}{Average normalized $\text{area}$ ($m^2$)}\\
		\midrule
		$R$ 	&	(same for all primitives)\hspace{3em}		            & 1.000\\
		$R_i$ 	&	(individual for every primitive)\hspace{3em}            & 0.646\\
		$R_i(t)$ &	(individual and time dependent)\hspace{3em}             & 0.329\\
		$p(x\vert t, \trajX^{i}),\;\forall i$&					&     \textbf{0.109}\\
		\midrule
	\end{tabular*}
	\caption*{Average normalized area (mean $\pm$ 2.6$\sigma$, equivalent to $\prob = 99\%$), normalized with respect to $R$ (using a single safety margin radius $R$ for all dimensions and all motion primitives).}
\end{table}
	
	\subsection{Abnormality detection}
	For the experiment, a total of 104 motion primitives were used. Apart from the 26 primitives in Fig. \ref{fig:motion-primitives} the other primitives have non-zero initial and/or final velocity. 
	For each primitive 20 randomized but different valid triplets were executed and recorded.
	The first 10 triplets for each primitive were used for learning the motion primitive model for each primitive, i.e. a training set.
	The following 10 were used for testing the abnormality detection, i.e. the test set.
	
	The naive approach of defining abnormal behavior as leaving the probability region is compared with the proposed approach using the Beta posterior.
	$\prob=99.9$\% is used.
	
	For the Beta-method the prior parameters is set to $\alpha = 1$ and $\beta = 999$.
	The posterior is calculated including on the observations from the last second, meaning we set the time window $W = 1 $ second.
	For each time point, we calculate the probability of the failure rate being larger than $(1 - \prob) = 0.01\%$.
	If the probability at any time point is larger than the threshold of 99.9\%, the entire execution is classified as abnormal, else normal.
	
	Both methods use the probability regions given from the learned motion primitive models derived from the training set and then tested on the test set.
	Figure \ref{confusion} shows the confusion matrices for each combination of method and data set.
	The Beta-method performs slightly better than the naive method, while still maintaining perfect precision.
	This suggests that it might be possible to tweak the 99.9\% threshold and the size of the time window to improve recall, without loss to precision.
	
	\newcommand\items{2}   %Number of classes
	\arrayrulecolor{white} %Table line colors
	\begin{figure}
		\vspace*{0.1in}
		\setlength{\extrarowheight}{0.15em}
		%\centering
		\begin{subfigure}[b]{0.4\columnwidth}
			\scalebox{0.78}{%
				\begin{tabular}{cc*{\items}{|E}|}
					\multicolumn{1}{c}{} &\multicolumn{1}{c}{} &\multicolumn{\items}{c}{Actual} \\ \hhline{~*\items{|-}|}
					\multicolumn{1}{c}{} & 
					\multicolumn{1}{c}{} & 
					\multicolumn{1}{c}{\rot{Normal}} & 
					\multicolumn{1}{c}{\rot{Abnormal}} \\
					\multirow{\items}{*}{\rotatebox{90}{Predicted}}
					&Normal    & 884  & 0  \\ \hhline{~*\items{|-}|}
					&Abnormal  & 0  & 91052 \\ \hhline{~*\items{|-}|}
				\end{tabular}
			}
			\vspace{0.5em}
			\captionsetup{justification=centering,margin=2cm}
			\caption{}
		\end{subfigure}~\hspace{1.5em}~
		\begin{subfigure}[b]{0.4\columnwidth}
			%\hspace{-0.2em}
			\scalebox{0.78}{%
				\begin{tabular}{cc*{\items}{|E}|}
					\multicolumn{1}{c}{} &\multicolumn{1}{c}{} &\multicolumn{\items}{c}{Actual} \\ \hhline{~*\items{|-}|}
					\multicolumn{1}{c}{} & 
					\multicolumn{1}{c}{} & 
					\multicolumn{1}{c}{\rot{Normal}} & 
					\multicolumn{1}{c}{\rot{Abnormal}} \\ \hhline{~*\items{|-}|}
					\multirow{\items}{*}{\rotatebox{90}{Predicted}}
					&Normal    & 884  & 0  \\ \hhline{~*\items{|-}|}
					&Abnormal  & 0  & 91052    \\ \hhline{~*\items{|-}|}
				\end{tabular}
			}
			\vspace{0.5em}
			\captionsetup{justification=centering,margin=2cm}
			\caption{}
		\end{subfigure}
		%\vspace{1em}
		\\
		
		\begin{subfigure}[b]{0.4\columnwidth}
			\scalebox{0.78}{%
				\begin{tabular}{cc*{\items}{|E}|}
					\multicolumn{1}{c}{} &\multicolumn{1}{c}{} &\multicolumn{\items}{c}{Actual} \\ \hhline{~*\items{|-}|}
					\multicolumn{1}{c}{} & 
					\multicolumn{1}{c}{} & 
					\multicolumn{1}{c}{\rot{Normal}} & 
					\multicolumn{1}{c}{\rot{Abnormal}} \\
					\multirow{\items}{*}{\rotatebox{90}{Predicted}}
					&Normal    & 711  & 0  \\ \hhline{~*\items{|-}|}
					&Abnormal  & 173  & 91052 \\ \hhline{~*\items{|-}|}
				\end{tabular}
			}
			\vspace{0.5em}
			\captionsetup{justification=centering,margin=2cm}
			\caption{}
		\end{subfigure}~\hspace{1.5em}~
		\begin{subfigure}[b]{0.4\columnwidth}
			%\hspace{-0.2em}
			\scalebox{0.78}{%
				\begin{tabular}{cc*{\items}{|E}|}
					\multicolumn{1}{c}{} &\multicolumn{1}{c}{} &\multicolumn{\items}{c}{Actual} \\ \hhline{~*\items{|-}|}
					\multicolumn{1}{c}{} & 
					\multicolumn{1}{c}{} & 
					\multicolumn{1}{c}{\rot{Normal}} & 
					\multicolumn{1}{c}{\rot{Abnormal}} \\ \hhline{~*\items{|-}|}
					\multirow{\items}{*}{\rotatebox{90}{Predicted}}
					&Normal    & 743  & 0  \\ \hhline{~*\items{|-}|}
					&Abnormal  & 141  & 91052    \\ \hhline{~*\items{|-}|}
				\end{tabular}
			}
			\vspace{0.5em}
			\captionsetup{justification=centering,margin=2cm}
			\caption{}
		\end{subfigure}
		\caption{Confusion matrix of abnormality detection over all primitives using a centered $99.9\%$-probability volume and (a, c) requiring that all observations fall inside this volume or (b, d) using a Beta posterior requiring the ratio of individual abnormalities less than $99.9\%$ probable to be above $0.01\%$ within a $1$s window. \textbf{First row:} Performance on the model training set. \textbf{Second~row:} Performance on previously unseen executions, from the same primitives, but different triplets.}
		\label{confusion}
	\end{figure}

	\addtolength{\textheight}{-0.55cm}	
	\section{CONCLUSIONS AND FUTURE WORK}
	\label{ch:conclusions}
	With increased autonomy of cyber-physical systems the need for integrated introspection capabilities is of growing importance. Such capabilities allow a robot to self monitor and to react to unexpected changes to circumstances in the environment. This is paramount if robots are supposed to operate safely in unstructured, dynamic and complex environments.
	We present an integrated approach for learning and monitoring the execution of motion actions, motion primitives, within the lattice-based motion planning paradigm.
	
	The feasibility is demonstrated empirically using a nonlinear dynamical model of a well known quadcopter platform in simulation. We observe that the motion primitives can be bad predictors of actual motion execution and that our learned models of executions drastically improve both the point predictive performance and the necessary safety margin is much reduced. Both of these aspects are important for effective collision checking during motion planning.
	
	Interesting future works is robust online learning of motion primitive execution models, as well as learning such models for different situations e.g. windy and non-windy conditions.
	
	Consecutive failures are currently modeled as independent of each other in the abnormality detection. Removing this assumption will likely improve the performance on false alarms.
	The presented approach fills an important role in robot safety and the presented work enhances modern lattice-based motion planning methods with effective collision avoidance and monitoring of model correctness, making robots increasingly reliable and safe in real-world settings.
	
	
	%%%%%%%%%%%%%%%%%%%%%%%%%%%%%%%%%%%%%%%%%%%%%%%%%%%%%%%%%%%%%%%%%%%%%%%%%%%%%%%%

	%%%%%%%%%%%%%%%%%%%%%%%%%%%%%%%%%%%%%%%%%%%%%%%%%%%%%%%%%%%%%%%%%%%%%%%%%%%%%%%%

	%%%%%%%%%%%%%%%%%%%%%%%%%%%%%%%%%%%%%%%%%%%%%%%%%%%%%%%%%%%%%%%%%%%%%%%%%%%%%%%%
	%\section*{APPENDIX}
	
	%\section*{ACKNOWLEDGMENT}

	%%%%%%%%%%%%%%%%%%%%%%%%%%%%%%%%%%%%%%%%%%%%%%%%%%%%%%%%%%%%%%%%%%%%%%%%%%%%%%%%
	\bibliographystyle{IEEEtran}
	\bibliography{bibliography}
	
\end{document}